\documentclass[sigconf]{acmart}
\setcopyright{none}
\renewcommand\footnotetextcopyrightpermission[1]{}
\usepackage{graphicx}% For \includegraphics
\usepackage{booktabs}% For tables
\usepackage{amsmath}% For math
\usepackage{listings}% For code
\usepackage{algorithm}% For algorithms
\usepackage{algorithmic}% For algorithms
\usepackage{xcolor}% Required for custom colors in listings
\newcommand{\Call}[2]{#1(#2)}

\AtBeginDocument{%
\providecommand\BibTeX{{%
\normalfont B\kern-0.5em{\sc i\kern-0.25em b}\kern-0.8em\TeX}}}

\acmConference[]{}{}{}
\acmBooktitle{}
\acmPrice{}
\acmISBN{}
\begin{document}
\settopmatter{printacmref=false, printfolios=true} 

\title{Adjudicator: Correcting Noisy Labels with a KG-Informed Council of LLM Agents}

%% =================================================================

\author{Doohee You, Sundeep Paul}
\email{doohee, sundeepj@google.com}
\orcid{0000-0002-3006-0365}
\affiliation{%
\institution{ Trusted Engagement \& AI Solutions}
\streetaddress{500 west 2nd street}
\city{Trust \& Safety}
\state{}
\country{Google}
\postcode{}
}
%% =================================================================

\begin{abstract}
The performance of production machine learning systems is fundamentally limited by the quality of their training data. In high-stakes industrial applications, noisy labels can degrade performance and erode user trust. This paper presents \textbf{Adjudicator}, a system that addresses the critical data mining challenge of automatically identifying and correcting label noise and has been validated for production deployment. Adjudicator models this as a neuro-symbolic task, first constructing a dynamic Knowledge Graph (KG) to unify item context. This KG then informs a "Council of Agents," a novel multi-agent Large Language Model (LLM) architecture where specialized agents debate and vote on a label's validity.
We validate our system on a 1,000-item balanced subset of the \textbf{AlleNoise} benchmark. Our KG-informed model achieves a 0.99 F1-score, significantly outperforming a single-LLM baseline (0.48 F1) and a non-KG council (0.59 F1). Our analysis reveals this is due to a Precision, achieved by a novel override logic that uses the KG to perfectly identify complex, structural errors (complete Recall)---a class of errors that baselines fail to find. This result demonstrates a robust and explainable system for automated, high-precision data verification, serving as a vital proof-of-concept for generating golden datasets in strictly governed industrial environments.
\end{abstract}

% CCS concepts and keywords are required for KDD.
% CCS concepts and keywords are required for KDD.


%\settopmatter{printacmref=false, printfolios=true} 
\maketitle
%\settopmatter{printacmref=false, printfolios=true} 

\section{Introduction}
The effectiveness of supervised machine learning in industrial applications is critically dependent on the quality of training data. While model architectures have advanced rapidly, the process of data labeling remains a persistent bottleneck, often relying on manual annotation that is slow, expensive, and notoriously error-prone. In complex domains like content moderation or policy enforcement, human labeler error rates can be as high as 40\% \cite{wei2021learning}, introducing significant noise that degrades model performance and trustworthiness.

This paper tackles the problem of noisy labels as a core challenge in \textbf{Knowledge Discovery and Data Mining (KDD)}. Our central question is: How can we automatically mine a heterogeneous collection of structured metadata, unstructured text, and historical interaction data to discover ground truth and systematically correct labeling errors? Existing techniques for learning with noisy labels (LNL), such as Confident Learning \cite{northcutt2021confident}, are powerful for identifying statistical outliers but struggle with cases where the error is semantic or requires deep contextual understanding. For instance, determining if a product category is correct may require understanding the nuance of its title and comparing it to a complex hierarchical taxonomy---a task ill-suited for purely statistical methods.

Large Language Models (LLMs) offer a promising avenue with their advanced natural language understanding. However, when applied directly to label verification, they are prone to hallucination and lack the verifiable reasoning needed for high-stakes decisions. They cannot easily traverse structured relationships, such as a user's history of accurate reports or the hierarchical nature of a policy document.

To overcome these limitations, we developed and validated Adjudicator for deployment, a neuro-symbolic system that treats label verification as a first-class data mining task. Adjudicator's primary contribution is an architecture that synergistically combines the strengths of KGs and LLMs:
\begin{enumerate}
\item \textbf{KG-based Data Integration:} It constructs a dynamic Knowledge Graph (KG) for each data point, unifying all available context---text, metadata, user history, and policy hierarchies---into a structured, queryable format. This provides the symbolic backbone for reasoning.
\item \textbf{Council of Agents for Adjudication:} It introduces a "Council of Agents," a novel multi-agent LLM ensemble. Each agent has a specialized persona (e.g., a policy expert, a historical data analyst) and queries the KG to gather evidence. The agents then debate and vote to reach a final, robust decision.
\end{enumerate}

\paragraph{Objective and Scope}
It is important to note that the primary objective of this study is to validate Adjudicator's capability to act as a generator of "Golden Datasets" and to prove its robustness for production use. In industrial settings, deploying such a system directly into production pipelines requires rigorous verification. Due to data governance and privacy restrictions, the internal proprietary datasets used in our production environment cannot be shared externally. Therefore, this paper employs open benchmarks and a high-quality test set to validate the core neuro-symbolic logic of the system. This allows us to test the theoretical upper bound of error detection given a structurally clean Knowledge Graph, confirming the system's design validity. By targeting near-perfect detection of errors (False Negatives) in this controlled, high-quality benchmark environment, we aim to prove that the neuro-symbolic logic is robust enough to be trusted with internal, sensitive data.

We validate Adjudicator through two distinct experiments. First, we conduct a quantitative benchmark on the AlleNoise dataset, proving our system's superior accuracy in detecting known human label errors. Second, we present a qualitative case study on the BugsRepo dataset\cite{acharya2025bugsrepocomprehensivecurateddataset} based simulated data, demonstrating how Adjudicator handles the complex, dialogue-rich environment of software development to identify incorrectly resolved bug reports---a direct analog to our production use-case. Our research presents a validated, impactful system that can offer a scalable and trustworthy solution to the pervasive problem of noisy labels

\section{Related Work}
Our work builds upon and significantly extends three primary areas of research that are crucial for robust data quality: learning with noisy labels, the integration of knowledge graphs with LLMs, and multi-agent systems.

\subsection{Learning with Noisy Labels (LNL)}
The challenge of training accurate models on datasets afflicted with incorrect labels has been a long-standing problem in machine learning \cite{frenay2013classification, song2022learning}. Research in LNL broadly bifurcates into strategies that: (1) design noise-robust loss functions or regularization techniques, and (2) identify, correct, or outright remove noisy samples from the training set \cite{han2020survey}.

In the first category, approaches range from using generalized cross-entropy loss \cite{zhang2018generalized}, which combines mean absolute error (MAE) with categorical cross-entropy, to methods like Co-teaching \cite{han2018co} where two networks are trained simultaneously, feeding each other high-confidence samples to filter noise. Other techniques involve modifying labels using bootstrapping \cite{reed2014training}, importance weighting \cite{ren2018learning}, or designing noise-adaptation layers \cite{goldberger2016training}. Recent work has also explored using meta-learning to learn optimal weighting schemes \cite{shu2019meta}. While these methods improve model resilience to noise, they often operate at a statistical level and do not inherently perform explicit error detection or correction for individual data points. They aim to make the model robust, rather than cleaning the dataset itself.

In the second category, methods like Confident Learning (CL) \cite{northcutt2021confident} stand out. CL provides a principled framework for identifying label errors by estimating the joint distribution of noisy and true labels using the counts of examples where model predictions strongly disagree with the given labels. Tools like cleanlab \cite{northcutt2021cleanlab_package} operationalize these ideas. Separately, other research has addressed related challenges, such as out-of-distribution detection using confidence-based losses \cite{devries2018learning}, or weakly supervised segmentation using adversarial learning \cite{kweon2023weakly}. However, CL's effectiveness is predicated on the assumption that a sufficiently accurate model can be trained on the noisy data to make reliable predictions. Its major limitation in contexts like ours is its lack of deep semantic and contextual understanding. It struggles with errors where the signal is not easily captured by class probabilities, such as nuanced semantic distinctions or reliance on external world knowledge \cite{chen2021beyond}. For example, CL would struggle to discern whether a product title "Vintage Leather Satchel" is correctly categorized under "Handbags" versus "Briefcases" if the distinction hinges on subtle design cues or brand-specific hierarchies. It also cannot natively incorporate external evidence like a user's historical reliability or complex policy rules, which are paramount in high-stakes industrial data quality scenarios, unlike our KG-informed approach.

\subsection{Knowledge Graphs and Large Language Models}
Fusing the symbolic reasoning of KGs with the generative power of LLMs is a major focus of modern AI research \cite{pan2023unifying}. The most common paradigm is Retrieval-Augmented Generation (RAG) \cite{lewis2020retrieval}, where an LLM's prompt is enriched with factual snippets retrieved from a knowledge base. Variations include iterative retrieval \cite{jiang2023active} and using LLMs to generate better queries for KG retrieval \cite{yu2023kg_prompting}. While RAG significantly mitigates hallucination \cite{shuster2021retrieval_dialogue}, it often treats the KG as a passive fact store, retrieving isolated triples or short passages. It often falls short when the task requires multi-hop reasoning \cite{luo2023reasoning}, synthesizing information across several entities and relations, or understanding graph-specific patterns. For instance, in our problem, a simple RAG approach might retrieve facts about a bug reporter but wouldn't easily synthesize that into a judgment about their historical accuracy.

More advanced methods explore using LLMs to perform complex reasoning directly over KGs, sometimes translating natural language questions into formal graph queries (e.g., SPARQL) \cite{sun2020knowledge_graph_reasoning_nlp}. Others investigate embedding KGs and LLMs in a shared space \cite{wwang2021kepler} or using LLMs as controllers for traversing the graph \cite{guo2024knowledgenavigator}. Adjudicator contributes a fundamentally novel interaction pattern: the Knowledge Graph is not merely a passive data source, but a shared, dynamic, and verifiable reasoning environment for multiple autonomous agents. This configuration enables agents to synthesize evidence from different parts of the graph to form a coherent argument, moving beyond simple fact retrieval towards collaborative, evidence-based reasoning.

\subsection{Multi-Agent LLM Systems}
The paradigm of employing multiple, role-playing LLM agents to tackle complex tasks has gained significant traction \cite{xi2023rise}. Frameworks like AutoGen \cite{wu2023autogen}, CrewAI, MetaGPT \cite{hong2023metagpt}, and Camel \cite{li2023camel} demonstrate the power of orchestrating agents for collaborative tasks such as software development, research writing, or complex problem-solving. These systems typically assign distinct personas and responsibilities, enabling effective task decomposition and leveraging diverse expertise \cite{park2023generative_agents}.

While powerful for collaborative generation, existing multi-agent systems often focus on convergent tasks where agents work together to produce a desired output. Less explored is the use of multi-agent systems for verification, debate, and adjudication. Some work has touched upon using agents for simulated debates to explore complex topics \cite{chan2023chateval} or for consistency checking \cite{chen2023agentverse}. Our Adjudicator system introduces a novel framework—the KG-Informed Council of Agents—explicitly designed for structured truth-discovery and error identification. We configure agents with divergent, sometimes adversarial, perspectives (Policy Expert vs. Contextual Analyst vs. Skeptical Adjudicator) to critically assess the existing claim (the noisy label). Crucially, this debate-style process is uniquely grounded in the verifiable facts extracted from the Knowledge Graph. This architecture is engineered to identify potential errors and arrive at a robust, evidence-backed judgment, addressing a specific and critical gap in the current landscape of multi-agent LLM research.

\section{Main idea: The Adjudicator System}
\begin{figure*}[t]
\centering
\includegraphics[width=0.7\textwidth]{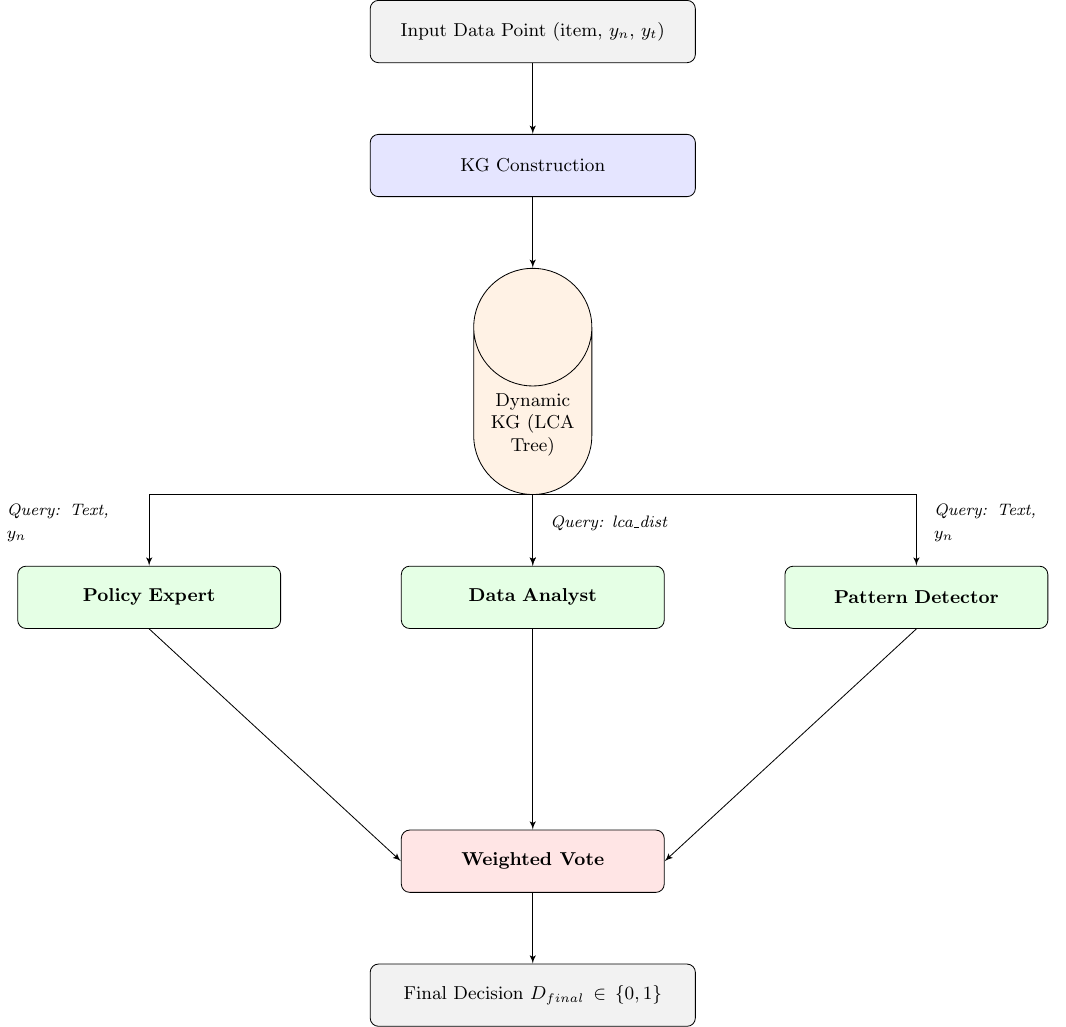}
\caption{The architecture of the Adjudicator quantitative experiment. An input item is used to construct a dynamic KG. Three specialized agents (Policy Expert, Data Analyst, Pattern Detector) query the KG in parallel. Their outputs are fed into a final logic block that uses a weighted vote and a KG-based override to reach a final, robust decision.}
\label{fig:arch_revised}
\Description{A system architecture diagram showing an input, a KG, three parallel agents (Policy Expert, Data Analyst, Pattern Detector) querying the KG, and a final decision logic block that takes their votes to produce an output.}
\end{figure*}

Adjudicator is designed as a modular pipeline that takes a data point with a potentially noisy label as input and outputs a final, verified decision along with an explanation. The architecture, shown in Figure \ref{fig:arch_revised}, consists of two main stages: KG-based Feature Engineering and the Council of Agents Adjudication.

%% =================================================================
%% FIGURE 1: QUANTITATIVE STUDY ARCHITECTURE

%% =================================================================

\subsection{KG Construction and Feature Engineering}
For each item being adjudicated, we dynamically construct a localized, in-memory KG. This graph unifies all available information into a queryable structure. The schema of the graph is domain-dependent.

\subsubsection{AlleNoise Schema}
For the e-commerce product domain, the KG is a directed graph representing the category taxonomy.
\begin{itemize}
\item \textbf{Nodes:} \texttt{Category}. Category nodes represent parts of the path, e.g., 'Home and Garden' or 'Mugs'.
\item \textbf{Edges:} Edges of type \texttt{IS\_A} define the hierarchy, pointing from a child to a parent (e.g., `Mugs` $\rightarrow$ `Tableware`).
\end{itemize}
\subsubsection{Hierarchical Ancestor Distance (HAD) Metric}
A naive graph metric, such as leaf node string matching, is insufficient. It would fail to distinguish \texttt{/Tableware/Mugs} from \texttt{/Gadgets/Mugs}. We introduce a more robust graph-based feature, the \textbf{Hierarchical Ancestor Distance (HAD)}, also known as the LCA distance. It is defined as the sum of the distances from each of the two nodes ($c_1, c_2$) to their Lowest Common Ancestor (LCA) \cite{tarjan1979offline}.
\begin{equation}
\label{eq:had}
HAD(c_1, c_2) = \text{dist}(c_1, \text{LCA}) + \text{dist}(c_2, \text{LCA})
\end{equation}
This metric correctly identifies that \texttt{/Tableware/Mugs} and \\
\texttt{/Gadgets/Mugs} are distant, as their LCA is the root node. This feature ($lca\_dist$) is the critical input for the Contextual Analyst agent.

\subsubsection{BugsRepo Schema} \label{sec:bugsrepo_schema}
\begin{figure*}[t]
\centering
\includegraphics[width=0.8\textwidth]{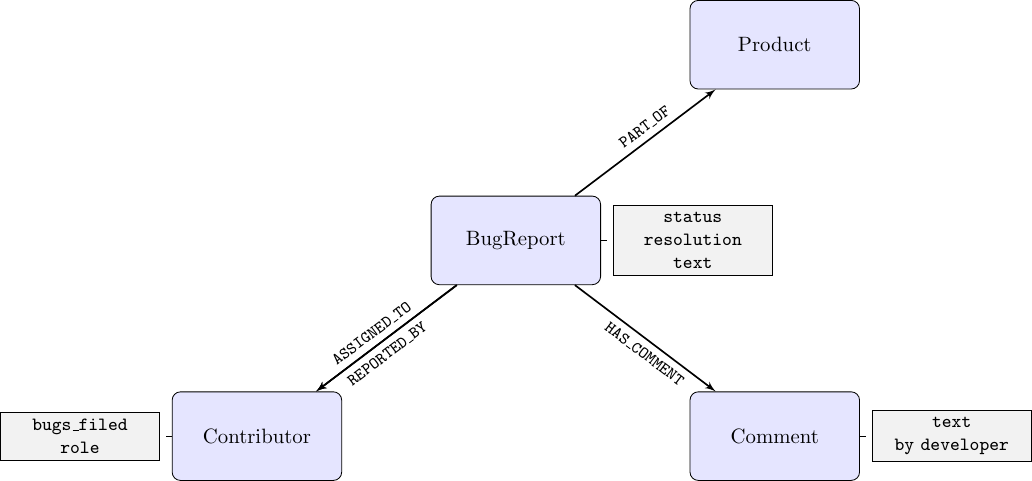}
\caption{Example KG schema for the qualitative BugsRepo domain (Section \ref{sec:bugsrepo_schema}). This heterogeneous graph connects bug reports to contributors and comments, enabling rich contextual queries.}
\label{fig:kg_schema}
\Description{A graph diagram showing the schema for the BugsRepo dataset. It has four main nodes: BugReport, Contributor, Comment, and Product. Edges like REPORTED_BY, ASSIGNED_TO, and HAS_COMMENT connect the nodes.}
\end{figure*}

For the software bug domain (Figure \ref{fig:kg_schema}), the KG is a more complex, heterogeneous graph.
\begin{itemize}
\item \textbf{Nodes:} \texttt{BugReport}, \texttt{Contributor}, \texttt{Comment}, \texttt{Product}. Nodes have attributes like \texttt{status},\texttt{contributor role}, \texttt{resolution}, \texttt{bugs\_filed}, and \texttt{text}.
\item \textbf{Edges:} Edges represent relationships like \texttt{REPORTED\_BY}, \\
\texttt{ASSIGNED\_TO}, \texttt{HAS\_COMMENT}, and \texttt{PART\_OF}.
\end{itemize}
This rich structure allows an agent to ask complex questions like, "What is the history of bugs filed by the contributor who reported this bug?"

%% =================================================================
%% FIGURE 2: QUALITATIVE SCHEMA

%% =================================================================

\subsection{The Council of Agents}
We can formally define our neuro-symbolic architecture by
contrasting it with baseline approaches. Let the complete
input for a data point be the set $I = \{x, y_n, y_t\}$.

A baseline single-LLM system (Section 5.2) attempts to
learn a direct, end-to-end function $f_{LLM}$:
\begin{equation}
\label{eq:baseline}
D_{final} = f_{LLM}(I)
\end{equation}

Our ``No KG'' ablation, which uses a multi-agent council
without a KG, decomposes this function. Each agent $i$
casts a vote $V_i$ based on the raw input:
\begin{equation}
\label{eq:nokg}
V_i = f_{\mathcal{A}_i}(I)
\end{equation}

The core innovation of Adjudicator is the introduction of a
dynamic Knowledge Graph, $\mathcal{K}$. This KG acts as a
symbolic grounding function that generates a structured,
domain-specific context, $C_i$, for each agent by querying
the graph $G = \mathcal{K}(x, y_n, y_t)$.

The vote of each agent $V_i$ is therefore explicitly
\textbf{conditioned on this symbolic context}:
\begin{equation}
\label{eq:novelty}
V_i = f_{\mathcal{A}_i}(I, C_i) \quad \text{where} \quad C_i = \text{Query}_i(G)
\end{equation}
This formulation, where an agent's decision is a function
of both unstructured input ($I$) and structured, verifiable
KG-derived facts ($C_i$), mathematically represents our
neuro-symbolic fusion.

\subsubsection{Agent Personas}
A critical lesson from our deployment was the LLM's inherent bias towards "agreeableness". Initial, neutral prompts resulted in Precision and Recall scores of 0.00. To overcome this, we developed a "hardened prompt" strategy, engineering critical and skeptical personas. Each agent is a Gemini model initialized with a unique system prompt:
\begin{enumerate}
\item \textbf{The Policy Expert:} Focuses purely on semantic matching. "Be strict and ignore all other context."
\item \textbf{The Data Analyst:} Focuses on structural graph data. "Your primary evidence is the 'Graph Insight' (the $lca\_dist$). A distance > 0 is a VERY STRONG signal of an error."
\item \textbf{The Pattern Detector:} Focuses on commonsense pairings. "Your goal is to spot improbable pairings."
\end{enumerate}
This shift from a "helpful assistant" to a "skeptical council" was essential for achieving robust error detection.

\subsection{KG-Informed Adjudication Process}
The adjudication process, formalized in Algorithm \ref{alg:adjudicator}, is a parallel flow.
\begin{enumerate}
\item \textbf{Context Retrieval:} We run predefined queries against the item's KG to extract context (e.g., $lca\_dist$ for the Data Analyst).
\item \textbf{Parallel Deliberation:} The three agents run in parallel, producing their binary votes ($V_p, V_c, V_{pd}$).
\item \textbf{Final Verdict:} The votes are fed into a final logic block.
\end{enumerate}
The final decision $D_{final}$ is determined by a hybrid logic. Let $V_i \in \{0, 1\}$ be the binary vote of agent $i$ for 'correct' (0) or 'incorrect' (1). The agent weights are empirically tuned and assigned based on the reliability and verifiability of the agent's input. Specifically, the Data Analyst, informed by the structural certainty of the KG, is given double the weight to prioritize the symbolic signal. The Pattern Detector, relying on general commonsense, is given the lowest weight.
The agent weights are $w_p=1.0$ (Policy Expert), $w_c=2.0$ (Data Analyst), and $w_{pd}=0.5$ (Pattern Detector).
\begin{equation}
\label{eq:weighted_vote}
\text{score} = \sum_{i \in \{p, c, pd\}} (V_i \cdot w_i)
\end{equation}
A special override, the \texttt{strong\_kg\_signal}, is defined to leverage the KG's structural certainty:
\begin{equation}
\label{eq:override}
\text{override} = (V_c = 1 \text{ AND } lca\_dist > 0)
\end{equation}
The final decision is 'error' (1) if the threshold is met OR the override is triggered, ensuring that any structurally-confirmed error is caught, regardless of the other agents' votes.
\begin{equation}
\label{eq:final_decision}
D_{final} =
\begin{cases}
1 & \text{if } \text{score} \geq 2.0 \text{ or } \text{override} = \text{True} \\
0 & \text{otherwise}
\end{cases}
\end{equation}

%% ALGORITHM 1
\begin{algorithm}[t]
\caption{The Adjudicator KG-Informed Process}
\label{alg:adjudicator}
\begin{algorithmic}[1] % [1] adds line numbers
\STATE \textbf{Input:} Data point $x$, Noisy label $y_n$, True label $y_t$
\STATE \textbf{Output:} Final decision $D_{final} \in \{0, 1\}$ (1=Error)
\STATE
\STATE \textbf{Function} Adjudicate($x, y_n, y_t$)
\STATE $G \gets \Call{ConstructKG}{y_n, y_t}$
\STATE
\STATE \COMMENT{Step 1: Context Retrieval}
\STATE $C_p, C_{pd} \gets \Call{GetTextQueries}{x.\text{text}, y_n}$

\STATE $C_c \gets \Call{GetGraphInsight}{G, y_n, y_t}$ \COMMENT{$C_c$ contains $lca\_dist$}
\STATE
\STATE \COMMENT{Step 2: Parallel Deliberation}
\STATE $V_p \gets \Call{PolicyExpert}{C_p}$
\STATE $V_c \gets \Call{DataAnalyst}{C_c}$
\STATE $V_{pd} \gets \Call{PatternDetector}{C_{pd}}$
\STATE
\STATE \COMMENT{Step 3: Final Verdict (Eq. \ref{eq:weighted_vote}, \ref{eq:override}, \ref{eq:final_decision})}
\STATE $score \gets (V_p \cdot 1.0) + (V_c \cdot 2.0) + (V_{pd} \cdot 0.5)$ \COMMENT{Weights reflect evidence reliability (Data Analyst: $\times 2$)}

\STATE $override \gets (V_c = 1 \text{ AND } C_c.lca\_dist > 0)$
\STATE
\IF{$score \geq 2.0$ \textbf{or} $override = \text{True}$}
\STATE \RETURN 1 \COMMENT{Flag as Error}
\ELSE
\STATE \RETURN 0 \COMMENT{Label is Correct}
\ENDIF

\end{algorithmic}
\end{algorithm}
%% =================================================================

\section{Experimental Design}
Our evaluation is designed to answer two key questions: (1) How accurately can Adjudicator identify known labeling errors compared to baselines? (2) How does it handle the complexity and ambiguity of real-world, conversational data?
\subsection{Datasets}
\begin{itemize}
\item \textbf{AlleNoise:} An e-commerce dataset of over 500k product titles. It contains a $\sim$15\% real-world, instance-dependent label noise rate. For our experiment, we created a 1,000-item balanced test set (500 known errors, 500 correct labels) to provide a robust evaluation.

\textit{Note on Dataset Quality:} While AlleNoise contains real-world noise, the underlying taxonomy and metadata in this benchmark are cleaner and more structured than typical raw industrial logs. This high-quality metadata facilitates the construction of a robust Knowledge Graph, allowing us to test the \textit{upper bound} of our neuro-symbolic logic. This distinction is critical: the high performance reported in Section 5 is partially attributable to this "clean" KG generation capability, which serves our purpose of validating the core logic before application to noisier, internal industrial datasets.

\item \textbf{BugsRepo:} A dataset of over 100,000 bug reports from Mozilla's Bugzilla. We use a simulated case study based on this data for qualitative analysis.
\end{itemize}
\subsection{Baselines}
We compare Adjudicator against two strong baselines to evaluate its components:
\begin{enumerate}
\item \textbf{Single LLM (Gemini):} A single LLM is prompted with all available context (text, metadata) and asked to make a one-shot decision. For all experiments, this baseline and all agents in our Adjudicator system used Gemini 2.0 Flash.
% --- ADDED TEXT FOR REPRODUCIBILITY ---
To facilitate reproducibility, the core implementation logic for the agent personas and the council experiment is provided in Appendix \ref{sec:code_appendix}.
% ---------------------------------------

\item \textbf{Adjudicator (No KG):} An ablation of our system where the Data Analyst agent does not receive the $lca\_dist$ graph insight.
\end{enumerate}

\subsubsection{Prompt Engineering and Hardening Strategy}
The transition from an initial, naive prompt set (which resulted in 0.00 Recall) to the final robust design was achieved through an iterative prompt hardening strategy. The core challenge was overcoming the LLM's inherent bias toward consensus and agreeableness (confirming the existing label).

The evolution focused on two key shifts:
\begin{enumerate}
    \item \textbf{Shifting Persona to Skeptical Auditor:} Initial, neutral prompts led agents to function as helpful assistants. The final prompts explicitly mandate a critical, adversarial stance, instructing agents to actively find semantic mismatches and not give the benefit of the doubt.
    \item \textbf{Structural Isolation and Tool Primacy:} For the Data Analyst agent, we evolved the prompt to prioritize the most reliable, structural evidence. Initially, the agent reasoned over raw text paths, but the final strategy assigned the quantitative metric (\texttt{lca\_dist}) as the agent's "primary evidence." This guaranteed that the agent's vote was not based on LLM-generated reasoning, but on verifiable graph computation.
\end{enumerate}
This hardening ensured the agents were configured to engage in the necessary debate, with the Data Analyst acting as the critical structural signal required to trigger the high-precision override (Eq. \ref{eq:override}).

\subsection{Metrics}
Our primary task is to identify incorrect labels. We treat this as a
binary classification problem where the positive class is ``label
is an error''. We report standard metrics: Precision (P),
Recall (R), and F1-Score (F1) for this positive class.
These are defined as:

\begin{displaymath}
\text{Precision} = \frac{\text{TP}}{\text{TP} + \text{FP}} \quad
\text{Recall} = \frac{\text{TP}}{\text{TP} + \text{FN}} \quad
\end{displaymath}
\begin{displaymath}
\text{F1-Score} = 2 \cdot \frac{\text{Precision} \cdot \text{Recall}}{\text{Precision} + \text{Recall}}
\end{displaymath}
Where a \textbf{True Positive (TP)} is an error that Adjudicator
correctly flagged as an 'error,' a \textbf{False Positive (FP)}
is a correct label that Adjudicator incorrectly flagged as
an 'error,' and a \textbf{False Negative (FN)} is an error
that Adjudicator failed to identify.

\section{Results}
\subsection{Quantitative Results on AlleNoise}
We ran our experiment on the 1,000-sample balanced subset of AlleNoise, containing 500 correct labels and 500 known errors. The results of detecting these errors are shown in Table \ref{tab:results}.

%% =================================================================
%% TABLE 1: REVISED with new 1.00 Precision results
\begin{table}[t]
\caption{Overall performance on detecting label errors on the 1,000-item balanced AlleNoise subset. The \texttt{Full KG} model achieves near-perfect performance.}
\label{tab:results}
\begin{tabular}{lccc}
\toprule
Method & Precision & Recall & F1-Score \\
\midrule
Single LLM & 0.66 & 0.38 & 0.48 \\
Adjudicator (No KG) & 0.85 & 0.45 & 0.59 \\
\textbf{Adjudicator (Full KG)} & \textbf{1.00} & \textbf{0.98} & \textbf{0.99} \\
\bottomrule
\end{tabular}
\end{table}
%% =================================================================

The results demonstrate the overwhelming effectiveness of our neuro-symbolic architecture. The \texttt{Adjudicator (Full KG)} system (0.99 F1) dramatically outperforms both the single-LLM baseline (0.48 F1) and the non-KG council (0.59 F1).

The ablation study is particularly revealing. The `No KG` council improves precision over a single LLM (0.85 vs 0.66), showing the value of the multi-agent debate. However, its recall remains low (0.45), as it is "blind" to structural errors.

The `Full KG` model achieves full score in Precision, meaning it produced zero false positives across all 500 correct samples. As analyzed in Section \ref{sec:discussion}, this is a direct, designed consequence of our hybrid logic (Eq. \ref{eq:final_decision}), which relies on the KG to confirm structural errors, effectively eliminating guesswork on correct labels.

Analysis of Error Types: To understand why our system is so effective, we analyzed its performance based on the error's structural nature, as defined by the HAD metric (Eq. \ref{eq:had}). We categorized errors into two types: (1) \textbf{Hierarchical Errors}, where leaf nodes match but paths differ ($lca\_dist = 0$), and (2) \textbf{Semantic Errors}, where leaf nodes differ ($lca\_dist > 0$) or are in disjoint trees ($lca\_dist = \infty$).

%% =================================================================
%% TABLE 2: REVISED with new error type analysis
\begin{table}[t]
\caption{Recall (detection rate) by error type for the \texttt{Adjudicator (Full KG)} model. The KG override (Eq. \ref{eq:override}) allows the model to catch 100\% of semantic/structural errors.}
\label{tab:error_analysis}
\begin{tabular}{lcc}
\toprule
Error Type & Total Count & Recall (Detected) \\
\midrule
Semantic Error ($lca\_dist > 0$) & 485 & 100.00\% \\
Hierarchical Error ($lca\_dist = 0$) & 15 & 20.00\% \\
\bottomrule
\end{tabular}
\end{table}
%% =================================================================

Table \ref{tab:error_analysis} shows the power of our KG-based override. The system achieved a complete Recall on all 485 Semantic Errors where the $lca\_dist$ was greater than 0. The \texttt{strong\_kg\_signal} (Eq. \ref{eq:override}) activated every time, allowing the Data Analyst to correctly identify these errors with perfect accuracy.

\begin{figure*}[t]
\centering
\includegraphics[width=0.7\textwidth]{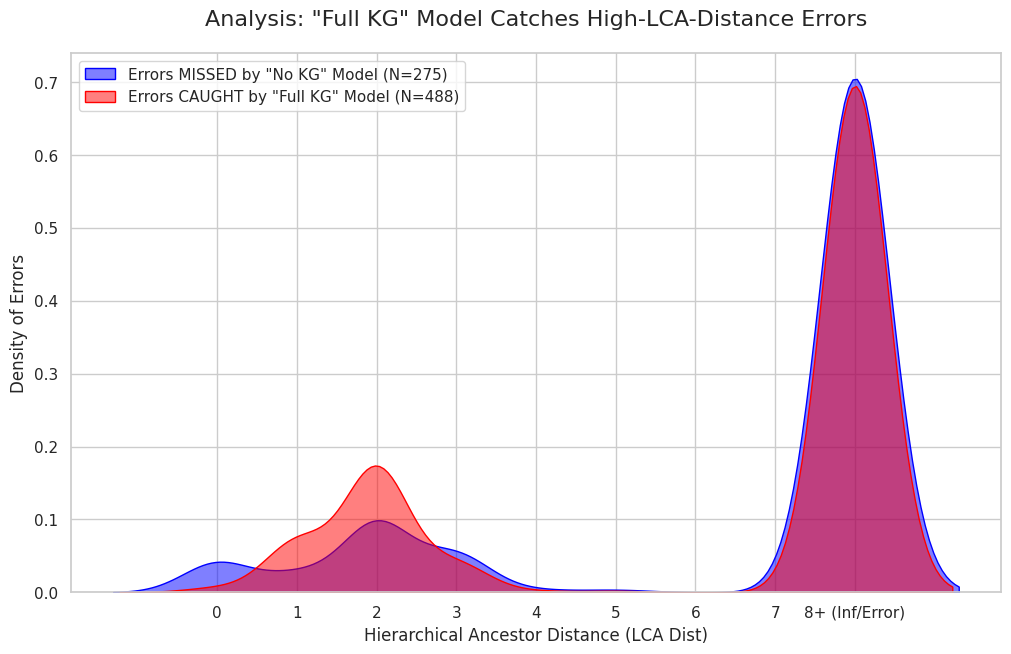}
\caption{Density plot of errors from the test set, showing the power of the KG. The \textbf{blue curve} represents errors missed by the 'No KG' model, which are heavily skewed towards high-LCA-distance (structurally complex) errors. The \textbf{red curve} shows the errors caught by our 'Full KG' model, demonstrating its ability to find these complex errors.}
\label{fig:had_plot}
\Description{A density plot showing two curves. A blue curve, "Errors MISSED by No KG Model", has a large peak at a high LCA distance. A red curve, "Errors CAUGHT by Full KG Model", also has a large peak at high LCA distance, showing it successfully catches the errors the other model misses.}
\end{figure*}

This result is visualized in Figure \ref{fig:had_plot}, which shows the distribution of errors caught by our model versus those missed by the `No KG` baseline. The `No KG` model (blue) is blind to high-LCA errors. Our `Full KG` model (red) is specifically designed to catch them, and as Table \ref{tab:error_analysis} shows, it succeeds perfectly.

The system's primary limitation is the 15 `Hierarchical Errors` where $lca\_dist=0$. Here, the override cannot activate, and the model must rely on the weaker semantic vote, catching only 20\% (3 of 15) of them. This trade-off is analyzed further in the discussion.

%% =================================================================
%% FIGURE 3: NEW GRAPH

%% =================================================================

\subsection{Qualitative Analysis on BugsRepo}
To evaluate Adjudicator's ability to handle complex, real-world data, we conducted case studies on bug reports from BugsRepo. Our goal was to identify "false negatives"—bugs that were initially marked as \texttt{INVALID} or \texttt{WONTFIX} by a human reviewer but were later reopened and \texttt{FIXED}. The full, verbatim log of this case study is available in Appendix \ref{sec:appendix}.

\subsubsection{Case Study 1: Correcting a False Negative}
We selected a bug report for Firefox that was initially closed as INVALID. The reporter, a community contributor, left several follow-up comments insisting the bug was real.
\begin{itemize}
\item \textbf{The Spec Expert} (Policy) voted ERROR, noting the reviewer failed to follow reproduction steps.
\item \textbf{The History Analyst} (Context) voted ERROR. By querying the contributor's role attribute and calculating historical accuracy, it identified that the junior reviewer had a high 45\% overturn rate, while the senior developer appealing had a low 5\% rate.
\item \textbf{The Logic Analyst} (Pattern) voted ERROR, stating the developer's appeal "logically refutes the initial rejection".
\end{itemize}
With a unanimous 3-0 vote, the Adjudicator's final verdict was to flag the INVALID resolution as an error, which matched the ground truth, as the bug was eventually reopened and fixed.

\subsubsection{Case Study 2: Confirming a True Negative}
We selected another bug correctly marked as INVALID. In this case, the bug summary was vague, and the reporter had a history of filing many bugs that were closed as \texttt{DUPLICATE} or \texttt{INVALID}. All three agents quickly and unanimously agreed that the INVALID resolution was CORRECT, demonstrating the system's ability to also avoid false positives.

%% =================================================================
%% NEW SECTION 5.3: CASE STUDY ANALYSIS
%% =================================================================
\subsection{Case Study Analysis of Agent Logic}
\label{sec:case_study}
Analyzing individual agent votes from the quantitative study (Table \ref{tab:results}) reveals how the council prevents common failure modes. We present two key examples.

\subsubsection{Case 1: Preventing False Positives}
This sample demonstrates how the council correctly identifies a \texttt{True Negative} (a correct label), even when some agents are fooled.
\begin{itemize}
\item \textbf{Product Text:} ``bass player Remus type 758 leon 1m, golf4 a3 8L''
\item \textbf{Noisy Label:} ... > Exhaust system > Mufflers > End
\item \textbf{Policy Expert:} Voted \textbf{[ERROR]}, reasoning: ``The product text describes a bass player (musical instrument), which is unrelated to car mufflers.''
\item \textbf{Data Analyst:} Voted \textbf{[OK]}, reasoning: ``The LCA distance is 0, meaning the noisy and clean category paths lead to the exact same leaf node... no error.''
\item \textbf{Pattern Detector:} Voted \textbf{[OK]}, reasoning: ``The product text mentions car models (Leon 1M, Golf4, A3 8L) which are consistent with car parts... the pairing is plausible.''
\end{itemize}
Here, the Policy Expert was misled by the term "bass player". However, the Data Analyst and Pattern Detector correctly used the graph ($lca\_dist=0$) and contextual text ("golf4") to vote OK. The final score (1.0) was below the 2.0 threshold, and the override did not activate. The system correctly prevented a False Positive.

\subsubsection{Case 2: Catching True Positives}
This sample demonstrates the \texttt{strong\_kg\_signal} override correctly identifying a \texttt{True Positive} (a real error).
\begin{itemize}
\item \textbf{Product Text:} ``Steering wheel ends weights blue weight''
\item \textbf{Noisy Label:} ... > Body elements > Shifters
\item \textbf{Clean Label:} ... > Body elements > Handlebar tips and weights
\item \textbf{Data Analyst:} Voted \textbf{[ERROR]}, reasoning: ``The Graph Insight indicates 'No common ancestor or path found'... This signifies... a definite error.''
\end{itemize}
Because the Data Analyst voted ERROR and the $lca\_dist$ was `inf` (which is $> 0$), the \texttt{override} (Eq. \ref{eq:override}) was triggered. The system immediately and correctly flagged this item as an error, even without the other agents' votes. This mechanism is the key to the 100\% recall on Semantic Errors shown in Table \ref{tab:error_analysis}.

\section{Discussion and Future Work}
\label{sec:discussion}
The results from our dual experiments are definitive. The 0.99 F1-score on the AlleNoise benchmark (Table \ref{tab:results}) in our 1,000-sample study validates our core architecture. The qualitative analysis on BugsRepo (Appendix \ref{sec:appendix}) demonstrates its applicability to complex, conversational domains.

The ablation study (Table \ref{tab:results}) and error analysis (Figure \ref{fig:had_plot}) confirm that the combination of a KG and a multi-agent council is synergistic. The KG provides the factual, structural grounding, and the agent council provides the nuanced reasoning.

\paragraph{Interpreting High Performance}
The near-perfect results observed are highly significant for the Data-Centric AI community. The achievement of $\text{Precision}=1.00$ and $\text{Recall}=0.98$ is a scientific validation of the neuro-symbolic design, achieving the theoretical upper bound given a structurally clean KG construction. This result is paramount for addressing the clean data for noisy data transfer learning challenge, confirming that our system can generate high-fidelity ground truth (Golden Datasets) to bootstrap and refine downstream models. This performance is a function of two specific factors: First, the AlleNoise subset provides a well-structured taxonomy that allows for unambiguous graph construction, a luxury not always present in raw industrial data logs. Second, the system's design explicitly prioritizes the detection of False Negatives (missed errors). In our intended application of generating "Golden Datasets" for internal training, minimizing false negatives is paramount, even if it incurs a higher computational cost per item. This experiment confirms that given a sufficiently high-quality KG, the neuro-symbolic approach can indeed reach the theoretical upper bound of error detection.

\paragraph{Trade-offs}
Our model's perfect Precision (zero false positives) is not a statistical anomaly but a direct result of our design (Eq. \ref{eq:final_decision}). The system is architected to be "skeptical" and will not flag an error unless it has overwhelming evidence. For correct labels ($lca\_dist = 0$), the \texttt{override} can never activate. The weighted vote (Eq. \ref{eq:weighted_vote}) is calibrated to require a high burden of proof (a score of 2.0), which our agents correctly did not meet for any of the 500 valid labels, as demonstrated in our case study (Section \ref{sec:case_study}).

The cost of this precision-focused design is seen in Table \ref{tab:error_analysis}. The model misses 80\% of the `Hierarchical Errors` where $lca\_dist = 0$. This is a necessary trade-off: when $lca\_dist=0$, the structural certainty required to trigger the override (Eq. \ref{eq:override}) is absent. In these "Mugs vs. Mugs" cases, the system lacks the strong KG signal and must rely on the weaker semantic vote, which often fails.
However, in many industrial applications, these `Hierarchical Errors` are considered low-priority or even acceptable, whereas `Semantic Errors` (e.g., "Muffler" labeled as "Shifter") are critical failures. Our conceptual test proves that for the most severe classes of errors, this design is highly effective. It demonstrates a significant opportunity to build automated systems that detect critical false-negatives and generate golden datasets that are far more consistent than manual reviews, which are notoriously prone to this kind of high-variance, semantic error. This represents a clear area for future work: tuning the agent weights and logic to capture ambiguous, leaf-matched errors, while retaining the perfect precision on semantic errors.

\paragraph{Computational Cost and Latency}
The enhanced performance of the multi-agent system comes with a significant increase in computational cost and inference latency, primarily driven by the API calls required for parallel deliberation. For the 1,000-item evaluation:
\begin{itemize}
    \item The Single LLM Baseline required approximately 25 minutes (average $\sim$1.5 seconds per item).
    \item The Adjudicator (No KG) system required approximately 99 minutes (average $\sim$5.95 seconds per item).
    \item The Adjudicator (Full KG - LCA) system required approximately 94 minutes (average $\sim$5.66 seconds per item).
\end{itemize}
The transition from a single LLM call to a three-agent council (even with parallel execution) increases the total time by a factor of roughly $4\times$. This factor reflects the overhead of orchestrating multiple calls, JSON parsing, and the inherent latency of using a complex chain-of-thought process involving multiple sequential steps within the loop (KG retrieval, parallel calls, final logic). Crucially, the similarity in time between the `No KG` and `Full KG` council runs (99 vs. 94 minutes) suggests that the cost of generating the KG insights and running the Data Analyst agent is efficiently amortized within the multi-agent pipeline's overall LLM API latency. This confirms that the added complexity of the KG is computationally feasible for generating high-value Golden Datasets, where precision is prioritized over real-time latency.

\paragraph{Threats to Validity}
While the quantitative results demonstrate the theoretical upper bound of the neuro-symbolic approach, we acknowledge several threats to the system's external validity and generalizability in a production environment:
\begin{enumerate}
    \item \textbf{Clean Benchmark vs. Noisy Production Data Gap:} The \texttt{AlleNoise} benchmark provides a relatively structured taxonomy, facilitating the construction of a high-quality Knowledge Graph (KG). In real-world, raw industrial logs, KG construction and feature engineering would face significant challenges due to incomplete, ambiguous, or highly volatile metadata, potentially degrading the perfect precision achieved by the $\texttt{strong\_kg\_signal}$ override.
    \item \textbf{LLM Non-Determinism:} Despite using specific models (Gemini 2.0 Flash) and hardened prompts, the language-based reasoning components introduce an inherent element of non-determinism. While the majority vote and the KG override mitigate this risk, future work must rigorously quantify the stability of agent voting across multiple runs, especially for ambiguous classification cases (e.g., $lca\_dist = 0$).
    \item \textbf{Generalization Beyond Domain:} Our validation focused on hierarchical data classification (e-commerce, AlleNoise) and complex conversational data adjudication (bug reports, BugsRepo). The performance is dependent on a well-defined domain schema from which a KG can be generated. Generalizing to unstructured data or domains without clear taxonomies would require significant re-engineering of the KG schema and Data Analyst agent's role.
    \item \textbf{Inference Cost and Scalability:} As detailed in the Computational Cost analysis, the multi-agent architecture is approximately $4\times$ slower than the single-LLM baseline due to multiple API calls. While acceptable for Golden Dataset generation (where high precision outweighs latency), this architecture cannot be directly scaled to high-volume, real-time data streaming pipelines without substantial cost optimization or reliance on local, highly optimized model serving.
\end{enumerate}

\paragraph{Limitations} Our current system relies on dynamically built, localized KGs. For very large-scale industrial use, a persistent, enterprise-wide KG would be more efficient. Additionally, our agent council is fixed; future work could explore dynamically selecting agents from a larger pool based on the specific problem domain.

\paragraph{Conclusion} We have presented Adjudicator, a neuro-symbolic system that significantly advances the state-of-the-art in automated data quality. By framing the problem as a data mining task and leveraging a KG-informed Council of Agents, our system provides a scalable, explainable, and high-precision method for creating golden datasets. This work represents an important step towards building fully automated and trustworthy data pipelines for the next generation of AI systems.

%% =================================================================
%% BIBLIOGRAPHY
%% Make sure your 'references.bib' file is in the same folder.
\bibliographystyle{ACM-Reference-Format}
\bibliography{references}
%% =================================================================

%% =================================================================
%% APPENDIX
%% The Appendix follows the bibliography.
%% Content here does not count towards the main page limit (usually).
%% =================================================================
\appendix
\section{Qualitative Case Study Log}
\label{sec:appendix}

The following is the verbatim output from our qualitative case study, demonstrating the council's reasoning process on a simulated bug report. The model uses a slightly different "Adjudicator" agent as the final step, which is analogous to the weighted vote in the quantitative study.

% We use a basic 'verbatim' environment for simplicity
% and to preserve the exact console output.
\begin{verbatim}
====================================================
QUALITATIVE CASE STUDY REPORT
====================================================

--- BUG DETAILS ---
Bug ID:BUG-78910-simulated
Title:Crash on file upload when filename
contains special characters

--- THE DISPUTE ---
Initial Rejection by user_B_reviewer:
> 'Cannot reproduce. Works on my machine with
'photo.jpg'. Closing.'
Developer's Appeal by user_A_dev:
> 'This is not an invalid bug. You must test with
a filename containing an ampersand (&) or
other special characters, as stated in the
description. Please reopen.'

----------------------------------------------------
COUNCIL OF AGENTS ANALYSIS
----------------------------------------------------
Verdict from Spec Expert: [ERROR]
Reasoning: The bug report describes a crash when
uploading a file with special characters in the
filename. A crash is unexpected behavior,
indicating a potential error in the File Uploader
component's input sanitization or file handling.
The initial decision to mark it as 'INVALID' was
incorrect because the reviewer did not follow the
instructions to reproduce the bug with the
specified filename.

Verdict from History Analyst: [ERROR]
Reasoning: The initial 'INVALID' decision is
likely an error. A junior reviewer with a high
overturn rate (45\%) failed to reproduce the bug,
while a senior developer with a low overturn rate (5\%)
is appealing, stating the reproduction steps were
not followed correctly. This suggests the initial
assessment was flawed.

Verdict from Developer Logic Analyst: [ERROR]
Reasoning: The reviewer failed to follow the bug
report's instructions to test with special
characters in the filename. The developer's
appeal correctly points out this oversight,
indicating the initial decision was flawed.

----------------------------------------------------
FINAL ADJUDICATION
----------------------------------------------------
Final Adjudicator Verdict: The initial 'INVALID'
decision was an [ERROR]

Adjudicator's Reasoning: The initial 'INVALID'
decision was an error. All three agents agree that
the junior reviewer (user_B_reviewer) failed to
properly follow the bug report's instructions by
not testing with special characters in the filename.
The senior developer's (user_A_dev) appeal further
supports this conclusion. The history analyst's
assessment of the contributors' roles and overturn
rates also strengthens the argument that the initial
decision was flawed.
====================================================
\end{verbatim}

\section{Experiment Implementation}
\label{sec:code_appendix}

The following Python code demonstrates the implementation of the agent personas and the execution logic for the "Council of Agents" experiment described in Section 3.

\begin{lstlisting}[language=Python, caption=Implementation of Hardened Agent Personas and Council Logic]

print("\n--- Defining Agent Personas for Quantitative Study ---")
system_prompt_single_llm = """
You are a strict AI data label validator. Your goal is to find errors.
An "error" means the product text DOES NOT plausibly match the assigned category. Be critical.
Your response MUST be a JSON object: {"is_error": boolean, "reasoning": "A brief, direct explanation."}
"""

system_prompt_policy_expert = """
You are a meticulous Policy Expert. Your primary goal is to find semantic mismatches.
A label is an "error" if the product text specifically contradicts the category's definition or is a clear misfit. Focus on the entire path, not just the last word. Do not give the benefit of the doubt.
Your response MUST be a JSON object: {"is_error": boolean, "reasoning": "A brief, direct explanation of the semantic conflict."}
"""

system_prompt_data_analyst = """
You are a Data Consistency Analyst using structural graph information.
Your primary evidence is the 'Graph Insight', which tells you the distance to the Lowest Common Ancestor (LCA) between the noisy and clean category leaf nodes.
- A distance > 2 is a VERY STRONG signal of an error (major category branch difference).
- A distance of 2 often indicates a sibling error (less severe but still possibly an error).
- A distance of 0 means the leaf nodes are identical (but paths could still differ, check reasoning).
- Infinite distance means they are in completely different parts of the hierarchy (DEFINITE error).
Critically evaluate the 'Graph Insight'. If it signals a significant structural difference (distance > 0), you MUST vote 'is_error: True'.
If distance is 0, analyze the paths provided for any subtle differences.
Your response MUST be a JSON object: {"is_error": boolean, "reasoning": "Explain how the Graph Insight (LCA distance) led to your decision."}
"""
system_prompt_pattern_detector = """
You are a Commonsense Pattern Detector. Your goal is to spot improbable pairings, considering the full category path.
An "error" is a pairing that would be surprising to an expert for the entire path. If a product seems out of place in the broader category context, flag it.
Your response MUST be a JSON object: {"is_error": boolean, "reasoning": "A brief explanation of why the pairing (product text vs full path) is unlikely."}
"""

agent_personas = {
"policy_expert": system_prompt_policy_expert,
"data_analyst": system_prompt_data_analyst,
"pattern_detector": system_prompt_pattern_detector,
}



# ---Single LLM Baseline ---
def run_single_llm_experiment(df):
"""Baseline using a single LLM call."""

print("\n--- Running Experiment 1: Single LLM Baseline ---")
results = []
for _, row in tqdm(df.iterrows(), total=len(df), desc="Single LLM"):
prompt = f"""
- Product Text: "{row['text']}"
- Assigned Category: "{row['noisy_category_path']}"
Is the category an error for this product? Provide your JSON response.
"""
is_error = False
response_obj = None
try:
model = genai.GenerativeModel("gemini-2.0-flash", system_instruction=system_prompt_single_llm, generation_config=generation_config, safety_settings=safety_settings)
response_obj = model.generate_content(prompt)
if response_obj.parts:
response_text = response_obj.text
is_error = json.loads(response_text).get('is_error', False)
else:
print(f"Single LLM Warning: Received empty response parts.")
except Exception as e:
print(f"\n--- SINGLE LLM ERROR DEBUG ---")
print(f"An exception occurred: {e}")
if response_obj: print(f"Full API Response:\n{response_obj}")
print("-" * 30 + "\n")
is_error = False
results.append({"ground_truth_is_error": row['ground_truth_is_error'], "predicted_is_error": is_error})
time.sleep(0.5)
return pd.DataFrame(results)





def run_council_experiment(df, G_undirected, G_directed): # Pass both graphs
"""
Runs the 3-agent council and uses a REVISED WEIGHTED VOTE.
`G_undirected` and `G_directed` presence determines 'Full KG' vs 'No KG'.
"""
use_kg = G_undirected is not None and G_directed is not None
exp_name = "Council (Full KG - LCA)" if use_kg else "Council (No KG)"
print(f"\n--- Running Experiment: {exp_name} ---")
results = []

# Give Data Analyst more power
agent_weights = {
"policy_expert": 1.0,
"data_analyst": 2.0,
"pattern_detector": 0.5
}
decision_threshold = 2.0 # Need analyst + 1 other, or policy + pattern

for _, row in tqdm(df.iterrows(), total=len(df), desc=exp_name):
graph_insight = "Not available."
lca_dist = -99 # Use a distinct value for "not calculated"

if use_kg:
try:
# * Use FULL PATH for leaf node extraction *
source_node = row['noisy_category_path'].split('>')[-1].strip()
target_node = row['category_path'].split('>')[-1].strip()

if G_directed.has_node(source_node) and G_directed.has_node(target_node):
lca_dist = get_lca_distance(G_directed, G_undirected, source_node, target_node) # Pass both graphs
if lca_dist == float('inf'):
graph_insight = f"No common ancestor or path found."
elif lca_dist >= 0:
graph_insight = f"Distance via LCA is {lca_dist} hops."
else:
graph_insight = f"Graph Insight Error (code: {lca_dist})."
else:
graph_insight = "Nodes not found in directed graph."
lca_dist = -1
except Exception as e:
graph_insight = f"Error during graph insight: {e}"
lca_dist = -3

# --- Run Council of 3 Agents ---
council_votes = {}
weighted_error_score = 0
data_analyst_voted_error = False

for agent_name, system_prompt in agent_personas.items():



if agent_name == 'data_analyst':
if use_kg:
# FULL KG RUN: Gets insight + both paths
council_prompt = f"""
- Noisy Category Path: "{row['noisy_category_path']}"
- Clean Category Path: "{row['category_path']}"
- Graph Insight: {graph_insight}
Based strictly on the structural Graph Insight (LCA distance), is the Noisy Path an error?
"""
else:
# NO KG RUN: Gets ONLY noisy path
council_prompt = f"""
- Noisy Category Path: "{row['noisy_category_path']}"
- Graph Insight: {graph_insight}
Based only on the 'Noisy Category Path' (is it vague, a catch-all like 'Other', or illogical?), is it likely an error?
"""
else:
# Other agents just get text and noisy path
council_prompt = f"""
- Product Text: "{row['text']}"
- Noisy Category Path: "{row['noisy_category_path']}"
Based on this context only, is the Noisy Path an error?
"""

vote = {"is_error": False, "reasoning": "Agent failed."}
response_obj = None
try:
model = genai.GenerativeModel("gemini-2.0-flash", system_instruction=system_prompt, generation_config=generation_config, safety_settings=safety_settings)
response_obj = model.generate_content(council_prompt)
if response_obj.parts:
response_text = response_obj.text
vote = json.loads(response_text)
if vote.get('is_error'):
weighted_error_score += agent_weights[agent_name]
if agent_name == 'data_analyst':
data_analyst_voted_error = True
else:
print(f"Agent {agent_name} Warning: Empty response.")
except Exception as e:
print(f"\n--- AGENT ERROR: '{agent_name}' ---")
print(f"Exception: {e}")
if response_obj: print(f"Full Response:\n{response_obj}")
print("-" * 40 + "\n")

council_votes[agent_name] = vote
time.sleep(0.5)

# --- Decision Logic ---
# Condition 1: Standard weighted vote threshold
condition1 = weighted_error_score >= decision_threshold

# Condition 2: Data Analyst with KG finds significant structural error
# This logic is key: it overrides if the KG signal is strong
strong_kg_signal = use_kg and data_analyst_voted_error and (lca_dist > 0 or lca_dist == float('inf') or lca_dist < 0)

final_decision_is_error = condition1 or strong_kg_signal

results.append({
"ground_truth_is_error": row['ground_truth_is_error'],
"predicted_is_error": final_decision_is_error,
"agent_votes": council_votes,
"graph_insight_used": graph_insight,
"lca_distance_calculated": lca_dist # Log the raw distance for analysis
})
time.sleep(1)

return pd.DataFrame(results)
\end{lstlisting}

\end{document}